# DCL-NET: DUAL CONTRASTIVE LEARNING NETWORK FOR SEMI-SUPERVISED MULTI-ORGAN SEGMENTATION

*Lu Wen[1,†], Zhenghao Feng[1,†], Yun Hou[2], Peng Wang[1], Xi Wu[3], Jiliu Zhou[1], Yan Wang[1,*]*

[1]College of Computer Science, Sichuan University, China.
[2]Southwest China Institute of Electronic Technology, Chengdu, China.
[3]College of Computer Science, Chengdu University of Information Technology, China.
[†]Co-first authors, contributed equally to this work.
[*]Corresponding author, email: wangyanscu@hotmail.com.

## ABSTRACT

Semi-supervised learning (SSL) is a sound measure to relieve the strict demand of abundant annotated datasets, especially for challenging multi-organ segmentation (MoS). However, most existing SSL methods predict pixels in a single image independently, ignoring the relations among images and categories. In this paper, we propose a two-stage Dual Contrastive Learning Network (DCL-Net) for semi-supervised MoS, which utilizes global and local contrastive learning to strengthen the relations among images and classes. Concretely, in Stage I, we develop a similarity-guided global contrastive learning to explore the implicit continuity and similarity among images and learn global context. Then, in Stage II, we present an organ-aware local contrastive learning to further attract the class representations. To ease the computation burden, we introduce a mask center computation algorithm to compress the category representations for local contrastive learning. Experiments conducted on the public 2017 ACDC dataset and an in-house RC-OARs dataset has demonstrated the superior performance of our method.

***Index Terms***— Semi-supervised learning, multi-organ segmentation, contrastive learning, deep learning

## 1. INTRODUCTION

Multi-organ segmentation (MoS) aims to simultaneously assign accurate category labels to pixels of multiple organs in the radiology images [1], playing an essential role in computer-aided treatment, such as disease diagnosis [2], radiotherapy [3, 4, 5], and survival prediction [6]. Recently, deep learning-based methods have gained impressive segmentation performance by fully-supervised training on large-scale labeled data [7]. Nonetheless, collecting such adequate labeled data is impractical in clinic for its expensive delineation cost.

To reduce the reliance on labeled data, semi-supervised learning (SSL) uses ample unlabeled and limited labeled data to jointly train the deep model to reach better performance. Current SSL methods can be divided into three groups: consistency regularization [8, 9, 10], proxy-label methods [11], and generative models [12]. Yet, these methods make predictions for pixels independently within a single image, ignoring the latent relations among images and classes.

To exploit the correlations among images or pixels, current works [13, 14, 15, 16] applied contrastive learning to medical image segmentation tasks and gained notable performances. Contrastive learning tries to force the feature representations of similar data to keep close while dissimilar ones to stay apart. Its key step is to effectively construct positive and negative pairs. For example, [13] proposed a global contrastive learning (GCL) to divide the volumes into several partitions and regarded the slices that belong to the same partition as positive pairs. Rather than rigid partitions, [14] introduced a positional contrastive learning (PCL) to use the relative position of slices and gains higher accuracy. Although these methods in a global fashion can provide the network with a global awareness, the local knowledge is also important for pixel-level dense prediction tasks like MoS. A common practice is to pull the feature representations of pixels with the same label together while pushing ones with different labels away [17]. But this formulation needs to calculate the similarity between every two pixel embeddings which may lead to heavy calculation burden.

In this paper, to address the aforementioned issues, we introduce a two-stage ***D**ual **C**ontrastive **L**earning **Net**work (DCL-Net)*, which incorporates both global and local contrastive learning to learn the discriminative representations for MoS. Specifically, in Stage I, we present a *similarity-guided global contrastive learning* strategy to pretrain a deep model to explore the implicit continuity and similarity among slices and learn richer global knowledge. In Stage II, we utilize the pretrained model to initialize a mean teacher model [8], and further train it with an *organ-aware local contrastive learning* strategy. Besides, a memory bank is maintained to store the representations of labeled data by organ categories. As complementary to the global contrastive learning, the local contrastive learning is conducted at the organ level to explore the relations among organ categories, thus encouraging the model to learn local representations highly related to organ segmentation.

Overall, the paper makes the following contributions: (1) We propose a novel dual contrastive learning model incorporating both global and local contrastive learning to excavate more comprehensive knowledge for the challenging semi-supervised MoS task. (2) To easy the computation burden, we design a mask center computation algorithm to compress the representations of the same category into a unique representation for the local contrastive learning. (3) Experiments conducted on the public 2017 ACDC dataset and an in-house rectum cancer organs-at-risk (RC-OARs) dataset have shown the superior performance of our method.

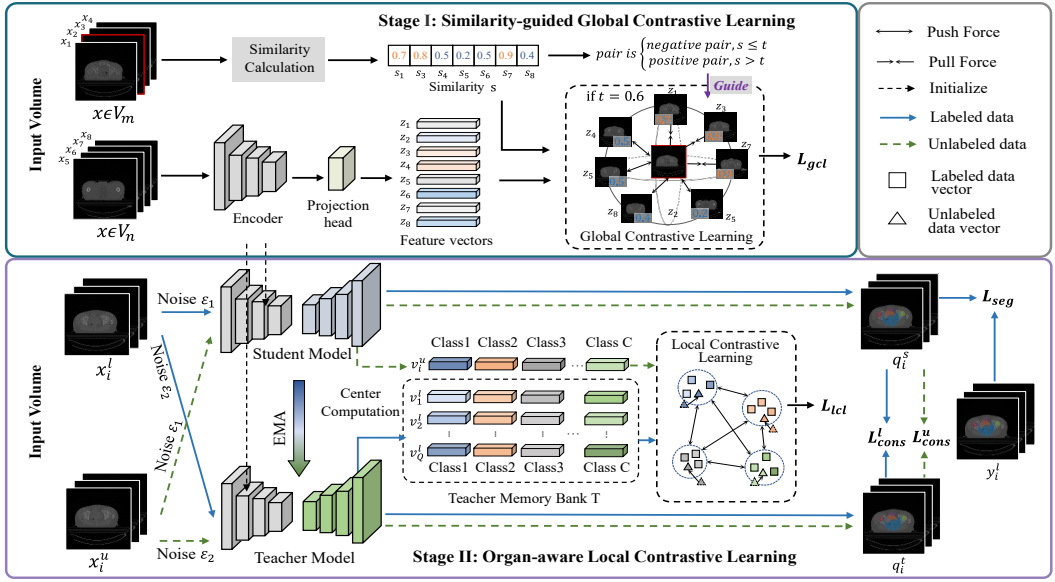

**Fig. 1.** Overview of the proposed DCL-Net.

## 2. METHODOLOGY

The overview of the proposed DCL-Net is displayed in Fig. 1, comprising two stages. In Stage I, fed with both labeled and unlabeled data, we calculate the slice similarities and pre-train an individual encoder with the global contrastive learning to learn global knowledge. Then, in Stage II, we build the segmentation network following the framework of mean teacher and use the well-trained encoder to initialize the encoders of student and teacher. Fed with both labeled and unlabeled data, mean teacher with different noise perturbations ($\varepsilon_1$ and $\varepsilon_2$) outputs the segmentation results, which are constrained by the consistency loss and supervised segmentation loss. To utilize the category information of different organs, we design an organ-aware local contrastive learning to learn more beneficial organ knowledge.

In our problem setting, the labeled set is defined as $D^l = \{x_i^l, y_i^l\}_{i=1}^N$ where $x_i \in R^{H \times W}$ is the 2D slice split from a 3D volume $V$, $y_i^l \in \{0,1 \dots C\}^{H \times W}$ is the segmentation label, and $C$ is the total class of substructures to be segmented. The unlabeled set is described as $D^u = \{x_i^u\}_{i=1}^M$ where $N \ll M$. More details will be described in subsequent sections.

### 2.1. Similarity-guided Global Contrastive Learning

We build the global contrastive learning based on the intrinsic slice continuity and similarity inside the medical volume. Concretely, we first process the 3D volume $V$ into $n$ 2D slices $\{x_i\}_{i \in [1,n]}$. Then, following the PCL [14], we utilize the relative positions ($p_i$ and $p_j$) of two slices ($x_i$ and $x_j$) in volumes to calculate the similarity $s_{ij}$ as below:

$$s_{ij} = |1 - (p_i - p_j)|. \quad (1)$$

Gaining 2D slices, we augment them and feed them into the feature extractor [14], including a UNet encoder $f(\cdot)$ [18] and a projection head $g(\cdot)$, and get the embedding vector $z_i = g(f(x_i))$. As slices $X = \{x_i\}_{i \in [1,B]}$ in a minibatch $B$ is assigned with position codes $Z = \{z_i\}_{i \in [1,B]}$, we can utilize the similarity $s_{ij}$ of two embeddings ($z_i$ and $z_j$) to construct the global contrastive loss as follows:

$$L_{gcl} = \sum_{i=1}^{2B} L_{gcl(i)},$$
$$L_{gcl(i)} = -\frac{1}{2B} \sum_{j=1}^{2B} s_{ij} \log \frac{\mathbb{1}_{[s_{ij}>t]} \exp(sim(z_i, z_j)/\tau)}{\sum_{k=1}^{2B} \mathbb{1}_{[i \neq k]} \exp(sim(z_i, z_k)/\tau)}, \quad (2)$$

where $sim(\cdot,\cdot)$ is the cosine similarity, and $\mathbb{1}_{[\cdot]}$ is a binary indicator, $t$ is the similarity threshold, $\tau$ is a temperature scaling parameter. Notably, different from treating all positive pairs fairly in PCL, we employ the similarity $s_{ij}$ to reweigh the loss value of different pairs, reducing semantically false positive pairs.

### 2.2. Organ-aware Local Contrastive Learning

To excavate the organ-level local information, we design a local contrastive learning based on mean teacher architecture in Stage II, which pulls/pushes the feature vectors from the same/different organ category of unlabeled and labeled data together/apart. We initialize the encoders of both student and teacher with the well-trained encoder $f(\cdot)$ in Stage I and train the segmentation decoders from scratch. The teacher model updates its parameters through exponential moving average (EMA) [8]. Fed with image $x_i$, the student and teacher output the prediction results, i.e., $q_i^s$ and $q_i^t$.

**Mask Center Computation.** To relive the computation burden of directly calculate the similarity between every two pixel embeddings, we propose a mask center computation strategy to compress all pixels in the same category into a center representation by masking and maintain it in the memory bank $T$. Specifically, the mask center representation $M_c$ of class $c$ can be expressed as an average of features of all pixels pertaining to class $c$:

$$M_c = \frac{\sum_{(i,j) \in L} \mathbb{1}_{[L_{(i,j)}=c]} p(e_{(i,j)})}{\sum_{(i,j) \in L} \mathbb{1}_{[L_{(i,j)}=c]}}, \quad (4)$$

where $e$ is the feature map generated by the second-to-last layer of the decoder, $p(\cdot)$ is a projection layer, $L$ is the mask, and $(i,j)$ denotes the pixel coordinate. For the labeled data, the mask $L$ is the real label, while for the unlabeled data, it is the pseudo label predicted by the segmentation network. Notably, feature $p(e)$ and mask $L$ have the same height and width, but $p(e)$ has $K$ channels and the dimension of $M_c$ is $1 \times K$. Therefore, the feature map $e$ of a single image $x_i$ can be compressed into a vector set $v_i = [M_c]_{c \in [1,C]}$, $v_i \in R^{C \times K}$. The vector sets of the whole labeled dataset is expressed as $V = \{v_i\}_{i \in [1,N]}, V \in R^{C \times K \times N}$.

**Teacher Memory Bank.** Considering the teacher updates its parameters through EMA and possesses more knowledge compared to the student, we allocate the memory bank to the teacher model. To ensure the quality of center representations in the teacher memory bank $T$, we only update $T$ with the latest $Q$ labeled images, i.e., a subset of $V$ produced by the teacher, denoted as $V' = \{v_i^l\}_{i \in [1,Q]}$. Hence, the dimension of bank $T$ is $C \times K \times Q$.

**Local Contrastive Learning.** We perform the local contrastive learning between the feature vector of unlabeled data, i.e., $v_i^u = [M_c^u]_{c \in [1,C]}$, and the stored vectors $V'$ in $T$ in the following formulation:

$$L_{lcl} = \sum_{c=1}^{C} L_{lcl}(c, M_c^u), \quad L_{lcl}(c, M_c^u) = -\frac{1}{|T_c|} \sum_{k_+ \in T_c} \log \frac{\exp(sim(M_c^u, k_+)/\tau)}{\exp(sim(M_c^u, k_+)/\tau) + \sum_{k_- \in [T \setminus T_c]} \exp(sim(M_c^u, k_-)/\tau)}, \quad (5)$$

where $k_+$ comes from the vector set $T_c$, and $k_-$ is from other vector sets not belonging to class $c$, i.e., $[T \setminus T_c]$. The loss pulls together the mask center representations of the same category of labeled and unlabeled data while pushing away those of different categories, thus learning features are more discriminative and sensitive to the pixel categories.

### 2.3. Objective Functions

The objective function of the whole network contains four parts: 1) global contrastive loss $L_{gcl}$, 2) local contrastive loss $L_{lcl}$, 3) supervised segmentation loss $L_{seg}$, and 4) consistency loss $L_{cons}$.

The supervised segmentation $L_{seg}$ can be expressed as:

$$L_{seg} = Dice(q_i^{s,l}, y_i^l) + CE(q_i^{s,l}, y_i^l). \quad (6)$$

Besides, the outputs of the student and teacher under different noise perturbations should keep consistent, so the consistency loss $L_{cons}$ can be formulated as follows:

$$L_{cons} = L_{cons}^l + L_{cons}^u,$$
$$L_{cons}^l = \sum_{i=1}^{N} \|q_i^{s,l} - q_i^{t,l}\|_2, \quad L_{cons}^u = \sum_{i=1}^{M} \|q_i^{s,u} - q_i^{t,u}\|_2. \quad (7)$$

Finally, the total loss function will be formulated as below:

$$L = L_{gcl} + \lambda_1 L_{lcl} + \lambda_2 L_{seg} + \lambda_3 L_{cons}, \quad (8)$$

where $\lambda_1$, $\lambda_2$, and $\lambda_3$ denote the weighting coefficients to balance the three loss terms.

## 3. EXPERIMENTS AND RESULTS

### 3.1 Dataset and Evaluations

We measure the performance of our model on two medical image segmentation datasets: (1) a public MRI dataset ACDC from MICCAI 2017 challenge [19], which includes 150 patients with segmentation labels of three organs: left ventricle, right ventricle, and myocardium; (2) an in-house CT dataset RC-OARs containing 130 rectum cancer patients with segmentation labels of four organs: bladder, small intestine, right femoral head, and left femoral head. We randomly select 80/20/50 and 100/8/22 samples as training/validation/testing sets for ACDC and RC-OARs dataset, respectively. Then, we split them into 2D slices for memory reduction and resize the resolution of slices as 352×352 for ACDC and 256×256 for RC-OARs. In the training set, we divide the labeled set and the unlabeled set as $n/m$ to simulate the semi-supervised setting, where $n$ and $m$ are the numbers of labeled and unlabeled samples. Besides, we employ the commonly used metrics, i.e., Dice coefficient and Jaccard Index (JI), to measure the accuracy, where the higher values represent the better performance.

### 3.2 Training Details

Our model is fulfilled with the PyTorch framework through a single NVIDIA GeForce 3090 GPU with 24GB memory. In Stage I, we use SGD optimizer to train the encoder for 200 epochs with batchsize of 64 and learning rate of 0.1. In Stage II, we use Adam optimizer to train the model for 100 epochs with batchsize of 16 and learning rate of 5e-4. The size of memory bank is set as 256. The local contrastive learning begins at the 30th epoch for stable convergence. In Stage I, the similarity threshold $t$ is set as 0.65 and 0.85 for the ACDC and RC-OARs dataset. $\lambda_1$ and $\lambda_2$ in Eq. (8) are empirically set as 0.01 and 1. As for $\lambda_3$, following [7], we use a warming-up function $\lambda_3(\delta) = 0.1 \times e^{(-5(1-\delta/\delta_{max})^2)}$ where $\delta$ and $\delta_{max}$ denote the current training step and total training steps.

### 3.3 Comparison with other SOTA methods

To verify the superior performance of our proposed model in MoS, we compare our method with six SOTA models, i.e., UNet (2015) [18], mean teacher (MT, 2017) [8], GCL (2020) [13], PCL (2021) [14], ICT (2022) [20], and RMT-VAT (2022) [21]. The quantitative results on ACDC dataset are listed in Table 1, where our method demonstrates a superior performance than other SOTAs in terms of all semi-supervised settings. Specifically, in the extreme case where only 5 labeled data are available, our method surpasses the baseline (UNet) largely by 20.3% Dice and 23.32% JI. As the number of label data increases, the proposed method still maintains the leading performance, beating the second-best performance by 6.33%, 3.01%, and 0.3%, respectively, in terms of Dice, when n=5, 10, and 20. Some visualization comparisons are illustrated in Fig. 2, from which we can intuitively see that the results of our method are closer to ground truth with less fault segmentation.

Table 1. Quantitative comparison with six SOTA methods on the ACDA dataset.

| $n/m$ | 5/75 | | 10/70 | | 20/60 | |
|---|---|---|---|---|---|---|
| | Dice | JI | Dice | JI | Dice | JI |
| UNet [18] | 59.87 | 44.04 | 77.79 | 64.43 | 86.52 | 76.50 |
| MT [8] | 70.42 | 55.18 | 80.12 | 67.38 | 88.27 | 79.23 |
| GCL [13] | 65.34 | 49.23 | 79.52 | 66.49 | 87.74 | 78.37 |
| PCL [14] | 66.08 | 50.17 | 80.93 | 68.65 | 88.70 | 79.90 |
| ICT [20] | 73.84 | 59.23 | 82.20 | 70.28 | 89.34 | 80.90 |
| RMT-VAT [21] | 72.25 | 57.63 | 83.59 | 72.33 | 89.91 | 81.87 |
| DCL-Net (Ours) | **80.17** | **67.36** | **86.60** | **76.64** | **90.21** | **82.33** |

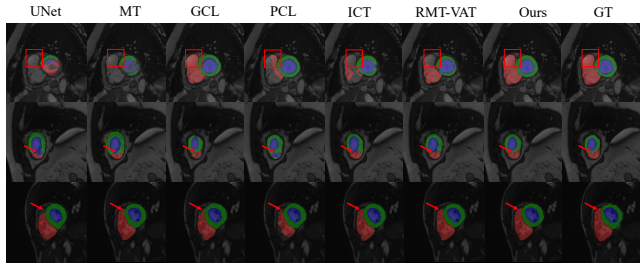

**Fig. 2.** Visualization comparisons on the ACDC dataset. From top to bottom, n=5, 10, and 20.

Furthermore, we also investigate the effectiveness of our method on RC-OARs dataset and the comparison results as displayed in Fig. 3. As observed, our DCL-Net obtains the best performance, i.e., 70.21% and 74.65%, when n=5 and 10, respectively, in terms of Dice. These results are 1.27% and 1.96% higher than the second-best method, i.e., RMT-VAT. Besides, our method also gains the least wrong segmentation in the visualizations. In summary, results on the two datasets have demonstrated the superior performance of our method both qualitatively and quantitatively.

### 3.4 Ablation study

To verify the effectiveness of important components of our DCL-MoS model, we conduct ablation experiments on the ACDC dataset. The experimental arrangements can be summarized as: (1) mean teacher as the backbone (MT), (2) MT + positional contrastive learning [14] (MT + $L_{pcl}$), (3) MT + similarity-guided global contrastive learning (MT + $L_{gcl}$), and (4) MT + $L_{gcl}$ + local contrastive learning (MT + $L_{gcl}$ + $L_{lcl}$) (DCL-Net, Ours). Quantitative results are given Table 2 where the MT model obtains the worst accuracy.

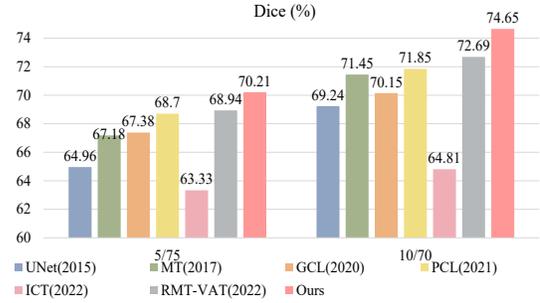

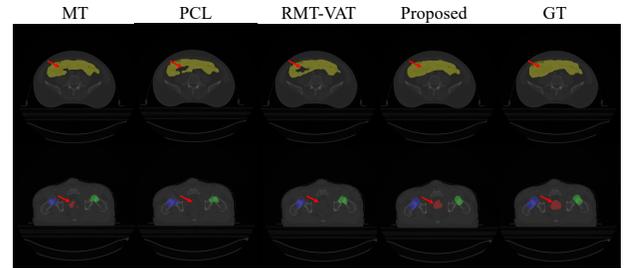

**Fig. 3.** The quantitative results on the RC-OARs dataset in terms of Dice are listed on the top. Corresponding visualization comparisons are shown on the bottom when n=5.

Table 2. Ablation study of our method on the ACDC dataset.

| $n/m$ | 5/75 | | 10/70 | | 20/60 | |
|---|---|---|---|---|---|---|
| | Dice | JI | Dice | JI | Dice | JI |
| (1) MT | 70.42 | 55.18 | 80.12 | 67.38 | 88.27 | 79.23 |
| (2) MT+$L_{pcl}$ | 73.02 | 58.87 | 82.51 | 70.68 | 89.19 | 80.71 |
| (3) MT+$L_{gcl}$ | 77.74 | 64.12 | 83.95 | 72.73 | 89.98 | 81.96 |
| (4) DCL-Net (Ours) | **80.17** | **67.36** | **86.60** | **76.64** | **90.21** | **82.33** |

After progressively performing the similarity-guided global contrastive learning and local contrastive learning, the model gains stable performance improvements even when the labeled data is relatively scarce, i.e., n=5, verifying their effectiveness. Notably, compared with the traditional PCL [14], our proposed similarity-guide contrastive learning improves the segmentation accuracy from 73.02% to 77.74% for Dice and from 58.87% to 64.12% for JI, respectively, when n=5, verifying its effectiveness in feature extraction.

### 4. CONCLUSION

In this paper, we propose a dual contrastive learning network DCL-Net to segment multiple organs in medical images under the semi-supervised scenario. The proposed method involves a similarity-guided global contrastive learning and an organ-aware local contrastive learning to extract rich feature expressions related to the MoS task. Experiments on the public ACDC dataset and the in-house RC-OARs dataset have shown the superior performance of our method.

**Acknowledgement.** This work is supported by National Natural Science Foundation of China (NSFC 62371325, 62071314), Sichuan Science and Technology Program 2023YFG0263, 2023YFG0025, 2023NSFSC0497.